\documentclass[sigconf]{acmart}

\AtBeginDocument{%
  }

\copyrightyear{2024}
\acmYear{2024}
\setcopyright{acmlicensed}
\acmConference[KDD '24] {Proceedings of the 30th ACM SIGKDD Conference on Knowledge Discovery and Data Mining }{August 25--29, 2024}{Barcelona, Spain.}
\acmBooktitle{Proceedings of the 30th ACM SIGKDD Conference on Knowledge Discovery and Data Mining (KDD '24), August 25--29, 2024, Barcelona, Spain}
\acmISBN{979-8-4007-0490-1/24/08}
\acmDOI{10.1145/3637528.3672352}

\usepackage{algorithm}
\usepackage{algpseudocode}
\usepackage{enumerate}

\usepackage{booktabs} 
\usepackage{multirow}

\usepackage{graphicx}
\usepackage{float} 
\usepackage{subfigure}

\begin{document}

\title{STATE: A Robust ATE Estimator of Heavy-Tailed Metrics for Variance Reduction in Online Controlled Experiments}

\author{Hao Zhou}
\authornote{Both authors contributed equally to this research.}
\orcid{0009-0008-5204-663X}
\affiliation{%
  \institution{State Key Laboratory for Novel Software Technology}
  \institution{Nanjing University}
  \city{Nanjing}
  \country{China}
}
\affiliation{%
  \institution{Meituan}
  \city{Beijing}
  \country{China}
}
\email{zhouhao29@meituan.com}

\author{Kun Sun}
\authornotemark[1]
\authornote{Corresponding author.}
\orcid{0009-0009-2282-8857}
\affiliation{%
  \institution{Meituan}
  \city{Beijing}
  \country{China}
}
\email{sunkun07@meituan.com}

\author{Shaoming Li}
\orcid{0000-0002-4915-9958}
\affiliation{%
  \institution{Meituan}
  \city{Beijing}
  \country{China}
}
\email{shaoming.li@outlook.com}

\author{Yangfeng Fan}
\orcid{0009-0001-8796-9396}
\affiliation{%
  \institution{Meituan}
  \city{Beijing}
  \country{China}
}
\email{fanyangfeng@meituan.com}

\author{Guibin Jiang}
\orcid{0009-0008-2558-4299}
\affiliation{%
  \institution{Meituan}
  \city{Beijing}
  \country{China}
}
\email{jiangguibin@meituan.com}

\author{Jiaqi Zheng}
\orcid{0000-0001-8403-9655}
\affiliation{%
  \institution{State Key Laboratory for Novel Software Technology}
  \institution{Nanjing University}
  \city{Nanjing}
  \country{China}
}
\email{jzheng@nju.edu.cn}

\author{Tao Li}
\orcid{0009-0004-3806-324X}
\affiliation{%
  \institution{Meituan}
  \city{Beijing}
  \country{China}
}
\email{litao19@meituan.com}

\renewcommand{\shortauthors}{Hao Zhou et al.}

\begin{abstract}

Online controlled experiments play a crucial role in enabling data-driven decisions across a wide range of companies. 
Variance reduction is an effective technique to improve the sensitivity of experiments, achieving higher statistical power while using fewer samples and shorter experimental periods.
However, typical variance reduction methods (e.g., regression-adjusted estimators) are built upon the intuitional assumption of Gaussian distributions and cannot properly characterize the real business metrics with heavy-tailed distributions.
Furthermore, outliers diminish the correlation between pre-experiment covariates and outcome metrics, greatly limiting the effectiveness of variance reduction.

In this paper, we develop a novel framework that integrates the Student's $t$-distribution with machine learning tools to fit heavy-tailed metrics and construct a robust average treatment effect estimator in online controlled experiments, which we call STATE. By adopting a variational EM method to optimize the loglikehood function, we can infer a robust solution that greatly eliminates the negative impact of outliers and achieves significant variance reduction. Moreover, we extend the STATE method from count metrics to ratio metrics by utilizing linear transformation that preserves unbiased estimation, whose variance reduction is more complex but less investigated in existing works. Finally, both simulations on synthetic data and long-term empirical results on Meituan experiment platform demonstrate the effectiveness of our method. Compared with the state-of-the-art estimators (CUPAC/MLRATE), STATE achieves over 50\% variance reduction, indicating it can reach the same statistical power with only half of the observations, or half the experimental duration.

\end{abstract}




\maketitle

\section{Introduction}

Online controlled experiments (also known as A/B tests) are the most widely adopted method for measuring causal effects, and play a crucial role in enabling data-driven decisions across a wide range of companies, including Facebook~\cite{Guo2021Machine,Bakshy2013Uncertainty}, Airbnb~\cite{Deng2023Variance,Deng2023Zero}, Google~\cite{Hohnhold2015Focusing}, Microsoft~\cite{Deng2013Improving} and LinkedIn~\cite{Ying2023Optimal,Xu2016Evaluating}. 
These experiments are critical for businesses, as even small differences detected in key metrics can have significant implications for the total revenue~\cite{Deng2013Improving}. 
For example, a strategy that increases one user's revenue by \$0.1 can result in millions of dollars in the total revenue for ten million users.


In the typical settings of controlled experiments, online traffic is randomly partitioned into two groups: a treatment group and a control group. They keep almost the same configuration except that the treatment group receives an additional intervention (e.g., a new production version or 
a promotion email). 
The average treatment effect (ATE) measures the causal effect of a treatment or intervention, 
but its groundtruth is unknown. Thanks to the central limit theorem, the difference-in-means estimator (DIM) produces an unbiased estimation of ATE, which is calculated by the difference between the treatment and control group outcomes.


Although DIM offers unbiased estimates, it still suffers from high variance, further leading to poor sensitivity and low statistical power. For online businesses, the sensitivity of experiments is particularly important. With thousands of experiments run each year, any benefits of increased sensitivity will be amplified due to economies of scale. On the other hand, the improvement in sensitivity allows experiments to be run on a smaller user population or for shorter durations while achieving the same statistical power.
It is significant for improving the product feedback cycle and agility.

In mathematics, variance reduction techniques are used to obtain higher precision for the metric of interest and have been introduced to online controlled experiments recently. 
For examples, CUPED~\cite{Deng2013Improving} utilizes pre-experiment covariates to reduce metric variability between the treatment and control groups, constructing an unbiased estimator with lower variance. 
However, the effectiveness is limited by the linear assumption and the correlation between the covariates and outcome metrics. Hence, to develop a nonlinear multi-covariate proxy that is highly correlated with the outcome metric, several machine learning (ML) based estimators have been explored~\cite{Aronow2013HT, Edwardwu2017loop, Ying2023Optimal, Guo2021Machine, Wager2016Highdimensional}. CUPAC~\cite{Huang2020Control} and MLRATE~\cite{Guo2021Machine} are two state-of-the-art (SOTA) estimators in this class, both of which are based on the regression-adjusted method.
To tackle the optimality, Yin~\cite{Ying2023Optimal} has demonstrated that if ML predicators converge to the conditional mean function, the asymptotic variance of ATE estimators can reach the semi-parametric variance lower bound.

\begin{figure}[htbp]
\centering  
\includegraphics[width=0.4\textwidth]{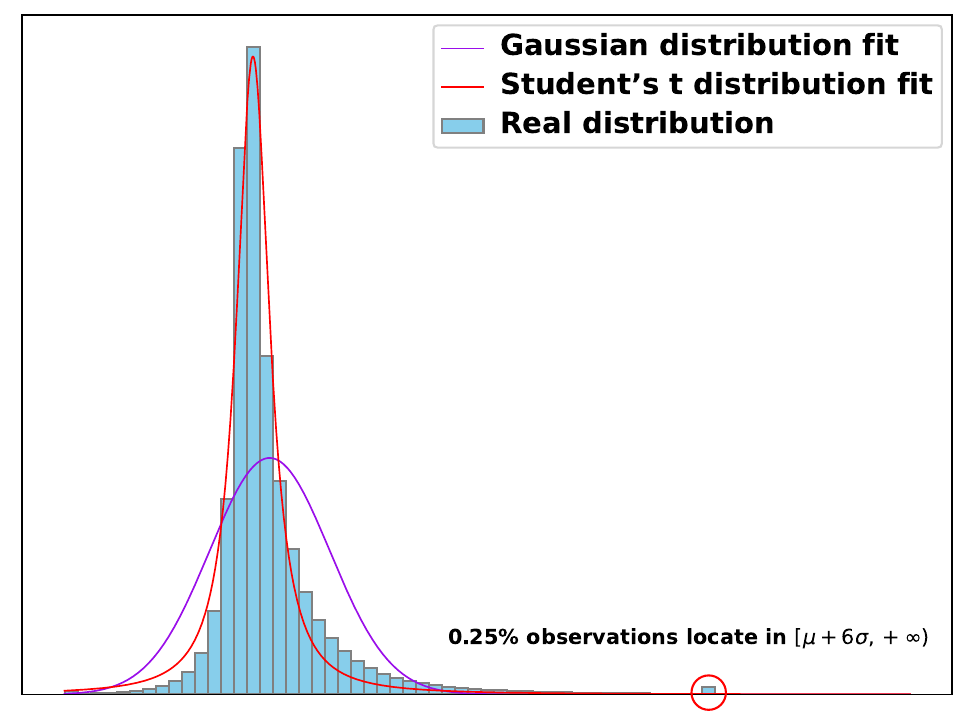}
\caption{The distribution of residuals for user GMV on the Meituan food delivery platform. The Y-axis represents the sample count. The X-axis represents the residuals between the real value and the model predicted value, 
where the extremely large values are clipped to the upper bound.
}
\label{fig:real-distribution}
\end{figure}

Notably, the discussion on optimality is conducted under the assumption that the regression residuals follow a Gaussian distribution. However the business metrics of interest are often heavy-tailed (e.g., watch time on video sites, user Gross Merchandise Volume(GMV) on e-commerce platforms, the amount of live broadcast rewards, etc.). The observations for these metrics usually contain some outliers (observations far away from the bulk of the probability density). The significant impact of outliers on the squared loss may lead to bias and high variance in the ATE estimates.
As shown in Fig.~\ref{fig:real-distribution}, the residuals of user GMV on the Meituan food delivery platform exhibit a heavy-tailed distribution due to the presence of users with extremely high-priced orders.  In the Gaussian distribution, the probability of observations falling beyond six standard deviations should be extremely small (less than 0.0000002\%). However, it is obvious that a larger portion of the observations exceeds six standard deviation in Fig.~\ref{fig:real-distribution}. Therefore, the Gaussian distribution does not properly characterize heavy-tailed metrics.

To address this problem, we introduce the Student's $t$-distribution to variance reduction. The Gaussian distribution is a special case of the $t$-distribution, corresponding to the situation where the degree of freedom tend to infinity. As the degree of freedom decreases, the $t$-distribution has heavier tails, giving non-zero probability to observations that fall outside the bulk of the density. This characteristic endows the $t$-distribution with an important property called robustness. As is shown in Fig.~\ref{fig:real-distribution}, 
the GMV metric can be better characterized when the regression residuals are modeled as a generalized $t$-distribution.


In this paper, we mainly focus on the variance reduction of ATE estimation of heavy-tailed metrics, and propose an easy-to-implement robust ATE estimator, called STATE. Specifically, our work has the following key contributions.

\begin{itemize}
    \item 
    %
    We integrate the machine learning regression adjustment method with the Student's $t$-distribution to estimate ATE for count metrics and derive a variational EM framework to infer parameters. The estimation procedure takes full advantage of both the powerful fitting ability of machine learning tools and the robustness of the $t$-distribution to outliers, thereby significantly reducing the variance of ATE estimation in online controlled experiments.


    \item We extend this method to ratio metrics, 
    which are more complex but less investigated. 
    We introduce linear transformation for ratio metrics while preserve unbiased estimation and consistent variance, under which regression-adjusted methods and the Student's $t$-distribution can be introduced to reduce variance and improve sensitivity.

    \item We conduct large-scale experiments on synthetic data and real business data on the Meituan food delivery platform to verify the effectiveness of our method in this paper. Extensive experimental results demonstrate that when the metrics follow Gaussian distribution, the STATE method performs equivalently to the state-of-the-art estimators (CUPAC/MLRATE).
    When the metric is a heavy-tailed distribution, our method can reduce the variance by about 50\% compared to the CUPAC and MLRATE. 
    This indicates that we can achieve the same statistical power in online controlled experiments with only half of the observations, or half the experimental duration.

\end{itemize}

\section{Setup and Related Work}
\subsection{Setup}
\label{subsec:setup}

Let $X$ be the pre-experiment covariates, $T$ be the binary treatment variable, and $Y,Z$ be two outcome variables.
In an online controlled experiment, suppose that there are $N$ samples denoted by $(X_i, T_i, Y_i, Z_i)$, which are drawn independently from an identical distribution. The treatment $T\in\{0,1\}$ is assigned randomly and independently of the covariates $X$, where $T_i=1$ represents that the $i$-th sample receives the treatment. Denote the number of samples with $T_i = 1$ and $T_i = 0$ by $N_t$ and $N_c$, repectively.
Following the Rubin Causal Model (RCM)~\cite{sekhon2008neyman}, let $Y_i(1), Z_i(1), Y_i(0), Z_i(0)$ be the corresponding potential outcome when the sample is assigned with the treatment or not. 

The count metrics are defined as the sample means (e.g., $\frac{\sum_{T_i=1}Y_i}{N_t})$, whose analysis unit is exactly the randomization unit in experiments. 
The average treatment effect (ATE) for count metrics is 
$$\tau_c = E[Y_i(1)] - E[Y_i(0)].$$
The common difference-in-mean (DIM) estimator is taken as 
\begin{equation} \label{Eq:Count-DIM}
\Delta Y = \frac{\sum_{T_i=1}Y_i}{N_t} - \frac{\sum_{T_i=0}Y_i}{N_c}.
\end{equation}
According to the central limit theorem, it gives an unbiased estimation of $\tau_c$ with the variance $D[\Delta Y] = \sigma_t^2/N_t+\sigma_c^2/N_c$, where $\sigma_t^2,\sigma_c^2$ are the variance of samples in the treatment group and control group respectively.

The ratio metrics are usually regarded as the ratio of two count metrics, e.g., $\frac{\sum_{T_i=1}Y_i}{\sum_{T_i=1}Z_i}$. For ratio metrics, ATE is defined as 
$$\tau_r = \frac{E[Y_i(1)]}{E[Z_i(1)]} - \frac{E[Y_i(0)]}{E[Z_i(0)]}.$$
The corresponding DIM estimator is 
\begin{equation} \label{Eq:Ratio-DIM}
\Delta R = \frac{\sum_{T_i=1}Y_i}{\sum_{T_i=1}Z_i} - \frac{\sum_{T_i=0}Y_i}{\sum_{T_i=0}Z_i}.
\end{equation}
Notice that DIM is not an unbiased estimator because
$$E[\Delta R] \neq \tau_r.$$
Fortunately, it gives a consistent estimation for $\tau_r$ according to the delta method~\cite{dasgupta2008asymptotic}.

\subsection{Related Work}
Variance reduction is a key technology and a longstanding challenge to improve the sensitivity of online control experiments. 
The original literatures mainly discuss the univariate linear adjustment method \cite{Yang2001Efficiency, Freedman2008Regression, Lin2013Agnostic, Deng2013Improving}. For instance, the CUPED \cite{Deng2013Improving} estimator, which is widely applied in industry, utilizes relevant covariate from the pre-experiment period to reduce the variability of the outcome metric, and shows that the higher correlation between covariate and outcome, the better the performance of variance reduction. Alternatively, an equivalent technique to CUPED is the linear regression method, $Y_i = a_0 + a_1T_i + a_2X_i + \epsilon_i$, which assumes the outcome is a linear combination of the treatment effect and the covariate term.

To overcome the limitations of the linear model, researchers have explored multivariate adjustment methods \cite{Guo2021Machine, Wager2016Highdimensional, Edwardwu2017loop,Aronow2013HT, Ying2023Optimal, Huang2020Control} by introducing the cross-fitting technique \cite{Chernozhukov2017Double, Zheng2011Cross} and “agnostic” regression \cite{Lin2013Agnostic, Freedman2008Regression}.
Typically, they utilize many covariates in a machine learning(ML) model $g$ predicting $Y$ from $X$ to develop a proxy variable $g(X)$.
Subsequently, estimate the ATE in a linear regression model $Y_i = a_0 + a_1T_i + a_2g(X_i) + \epsilon_i$.
Since the proxy incorporates more prior information about the outcome, it generates further variance reduction gains.



However, the regression-adjusted methods assume that the residual $\epsilon$ follows a Gaussian distribution, which is not applicable to the heavy-tailed metrics in online businesses. If outliers are not properly dealt with, they may lead to bias and high variance in parameter estimation.

Furthermore, the majority of variance reduction methods focus on count metrics. In fact, ratio metrics are equally important in practice, and are more complex but less investigated. Existing solutions for ratio metric are mostly extensions to those of count metrics. For examples, researchers utilize the delta method to extend the CUPED estimator to the variance reduction of ratio metrics. 
The study conducted by ~\cite{Ying2023Optimal} focuses on ratio metrics by separately minimizing the variances of the numerator and denominator, but disregards their correlation. 
A novel work proposed in ~\cite{Budylin2018Consistent} is the consistent transformation, which transforms ratio metrics into user-level linear metrics. However, it is based on a strong hypothesis where the denominator in ratio metrics is extremely large and can be approximatively taken as a constant.

In this paper, we develop more efficient estimators for count metrics and ratio metrics respectively. For count metrics, we propose a robust estimator called STATE by modeling the regression residual as a Student's $t$-distribution which was put forth and adopted as a robust building block, for clustering \cite{Peel2000Robust,Svensen2005Robust} and robust projections\cite{Archambeau2006Robust}.
The heavy-tailed nature of the $t$-distribution makes it much less sensitive to outliers compared to the Gaussian distribution, resulting in a substantial reduction in the variance of the ATE estimate. For ratio metrics, we adopt the main idea of the transformation method ~\cite{Budylin2018Consistent} while relaxing its assumptions, and introduce STATE method for ratio metrics to decrease the negative influence of outliers.

\section{Robust Estimation for Count Metrics with STATE}
\subsection{Robust Modeling}
Following the framework of typical regression adjustment techniques, our proposed ATE estimation procedure can be summarized in two stages: machine learning stage and linear regression stage. \\
\textbf{Machine learning stage:} utilize machine learning tools $g$ to capture the relationship between the outcome metric $Y$ and the covariates $X$. 
Then, we can obtain a proxy variable $\hat{Y}_i$ \cite{Guo2021Machine, Huang2020Control} composed of ML predictions $g(x_i)$.
Notably, we employ the cross-fitting technique \cite{Chernozhukov2017Double, Zheng2011Cross} here to ensure that the proxy $\hat{Y}_i$ and treatment $T_i$ are independent.\\
\textbf{Linear regression stage:} include the proxy $\hat{Y}_i$ as a regressor in the linear regression step. The ATE estimation is the estimate of $a_1$. 
\begin{equation}
\begin{aligned}
  Y_i &= a_0 + a_1T_i + a_2\hat{Y}_i + \epsilon_i   \label{eq:regression}
\end{aligned}
\end{equation}
for simplicity, we denote the Eq.~\eqref{eq:regression} as $Y_i = a^Tx_i + \epsilon_i$, where $a = (a_0,a_1,a_2)^T$, and $x_i= (1,T_i, \hat{Y}_i)^T$. \\

To capture the structure of the typical observations while dealing with outliers automatically, we model the residuals using the Student's $t$-distribution instead of the Gaussian distribution, here.
\begin{equation}
\epsilon_i \sim S_t(\epsilon|u,\sigma^2,v) \label{eq:noise}
\end{equation}

As noted in \cite{Liu2010ML}, the $t$-distribution can be re-written as a convolution of a Gaussian distribution with a Gamma distribution placed on its precisions by introducing a latent variable $\eta_i$.
\begin{equation}
S_t(\epsilon|u,\sigma^2,v)=\int_{0}^{\infty} \mathcal N(\epsilon|u,\frac{\sigma^2}{\eta_i}) \mathcal{G}(\eta_i|\frac{v}{2},\frac{v}{2})d\eta_i
\end{equation}
without loss of generality, the expectation $u$ of the residuals is set to 0. $\mathcal{G}$ is the Gamma density, $\mathcal{G}(\eta|a,b) = \frac{b^a}{\Gamma(a)}\eta^{a-1}e^{-b\eta}$, and $\Gamma(\cdot)$ represents the Gamma function.

It should be noted that the maximum likelihood estimate of $a_1$ based on Eq.~\eqref{eq:regression},\eqref{eq:noise} is the ATE estimation. Now we derive the log-likelihood function of the observations in the following form.
\begin{equation}
\begin{aligned}
\mathcal{L} &= \ln{\prod_{i=1}^{N} P(Y_i|x_i,\theta)}\\
    &= \sum_{i=1}^{N} \ln{S_t(Y_i|a^Tx_i,\sigma^2,v)}\\
    &= \sum_{i=1}^{N} \ln{\int_{0}^{\infty} \mathcal N(Y_i|a^Tx_i,\frac{\sigma^2}{\eta_i}) \mathcal{G}(\eta_i|\frac{v}{2},\frac{v}{2})d\eta_i}\\
    &\geq \sum_{i=1}^{N} \int_{0}^{\infty} q(\eta_i)\ln \frac{\mathcal N(Y_i|a^Tx_i,\frac{\sigma^2}{\eta_i}) \mathcal{G}(\eta_i|\frac{v}{2},\frac{v}{2})}{q(\eta_i)} d\eta_i\\
    &\equiv \mathcal{F} (Y|x,\theta,q) \label{eq:Likelihood}
\end{aligned}
\end{equation}
where $q(\eta_i)$ is the posterior of latent variable $\eta_i$, $\theta=\{a^T,\sigma^2,v\}$ is the set of parameters of the model. $\mathcal{F}$ is called the variational free energy function. The inequality holds due to Jensen Inequality.

\subsection{The Variational Likelihood Bound}
 The variational free energy $\mathcal{F}$, which is also called the Evidence Lower Bound(ELBO) of the likelihood function. Optimizing the likelihood function $\mathcal{L}$ directly is intractable, we can optimize the ELBO instead by the EM algorithm according to Minorize-Maximization optimization\cite{Lange2000Optimization,Parizi2015Generalized}. 
$\mathcal{F}$ can be evaluated as follows:
\begin{equation}
\mathcal{F} (Y|x,\theta,q) = \sum_{i=1}^{N} (\langle \ln{\mathcal{N}(Y_i|a^Tx_i,\frac{\sigma^2}{\eta_i})\rangle} + \langle \ln{\mathcal{G}(\eta_i|\frac{v}{2},\frac{v}{2})} \rangle - \langle \ln{q(\eta_i)} \rangle) \label{eq:Free_energy}
\end{equation}
where 
\begin{equation}
\langle \ln{\mathcal{N}(Y_i|a^Tx_i,\frac{\sigma^2}{\eta_i})\rangle} = -\frac{1}{2} \ln{2 \pi \sigma^2} + \frac{1}{2} \langle \ln \eta_i \rangle - \frac{\langle \eta_i \rangle}{2\sigma^2}(Y_i - a^Tx_i)^2  \label{eq:q1}
\end{equation}

\begin{equation}
\langle \ln{\mathcal{G}(\eta_i|\frac{v}{2},\frac{v}{2})} \rangle = \frac{v}{2} \ln{\frac{v}{2}} - \ln{\Gamma(\frac{v}{2})} + (\frac{v}{2} - 1) \langle \ln{\eta_i} \rangle - \frac{v}{2}\langle \eta_i \rangle \label{eq:q2}
\end{equation}

\begin{equation}
\langle \ln{q(\eta_i)} \rangle = \xi_i \ln{\zeta_i} - \ln{\Gamma(\xi_i)} + (\xi_i - 1)\langle \ln{\eta_i} \rangle - \xi_i \label{eq:q3}
\end{equation}
where $\langle \cdot \rangle$ represents the expectation of the term conditional on the posterior distribution $q(\eta_i)$, $\langle \ln{\eta_i} \rangle = \Psi(\xi_i) - \ln{\zeta_i} $ and $\Psi(\cdot)$ is the di-gamma function.

In the E-step of the $(k+1)$-th iteration of EM algorithm, we maximize $\mathcal{F}$ w.r.t the variational distribution $q$ while fixing the parameters in the $k$-th iteration, $\theta^k$:
\begin{displaymath}
q^{k+1}(\eta_i)=arg max_{q} \mathcal{F}(Y_i|x_i,\theta^k,q)
\end{displaymath}

In the M-step, we maximize $\mathcal{F}$ w.r.t the parameters $\theta$ to obtain the new parameter values $\theta^{k+1}$
\begin{displaymath}
\theta^{k+1}=arg max_{\theta} \mathcal{F}(Y_i|x_i,\theta,q^{k+1})
\end{displaymath}

\subsection{Deriving the EM Algorithm}
\subsubsection{Variational E-step}
\

\vspace{1mm}
The posterior distribution of latent variable $\eta_i$ can be derived by taking 
functional derivative of $\mathcal{F}(Y_i|x_i,\theta,q(\eta_i))$ w.r.t. the term of~$q(\eta_i)$,
\begin{displaymath}
q(\eta_i) = \frac{exp\langle \ln \mathcal N(Y_i|a^Tx_i,\frac{\sigma^2}{\eta_i}) \mathcal{G}(\eta_i|\frac{v}{2},\frac{v}{2}) \rangle}{\int_{0}^{\infty} exp\langle \ln \mathcal N(Y_i|a^Tx_i,\frac{\sigma^2}{\eta_i}) \mathcal{G}(\eta_i|\frac{v}{2},\frac{v}{2})\rangle d\eta_i }
\end{displaymath}
After simplification, we can find ~$q(\eta_i)$ is Gamma density.

\begin{displaymath}
q(\eta_i) = \mathcal{G}(\eta_i|\xi_i,\zeta_i)
\end{displaymath}
where 
\begin{equation}
\xi_i = \frac{v}{2}+\frac{1}{2}, \zeta_i=\frac{v}{2} +\frac{(Y_i-a^Tx_i)^2}{2\sigma^2},\langle \eta_i \rangle = \frac{\xi_i}{\zeta_i} \label{eq:posterior}
\end{equation}

\subsubsection{M-step}
\

\vspace{1mm}
The parameter $\theta=\{a^T,\sigma^2,v\}$ are obtained by solving the stationary equations of $\mathcal{F}$ w.r.t $a^T=\{a_0,a_1,a_2\}$, $\sigma^2$, $v$ respectively, 
\begin{equation}
a_0 = \frac{\sum_{i=1}^{N} (Y_i - a_1T_i - a_2\hat{Y}_i) \langle \eta_i\rangle}{\sum_{i=1}^{N}\langle \eta_i \rangle} \label{eq:a0}
\end{equation}

\begin{equation}
a_1 = \frac{\sum_{i=1}^{N} (Y_i - a_0 - a_2\hat{Y}_i)T_i \langle \eta_i\rangle}{\sum_{i=1}^{N}\langle \eta_i \rangle T_i^2} \label{eq:a1}
\end{equation}

\begin{equation}
a_2 = \frac{\sum_{i=1}^{N} (Y_i - a_0 - a_1T_i)\hat{Y}_i \langle \eta_i\rangle}{\sum_{i=1}^{N}\langle \eta_i \rangle \hat{Y}_i^2} \label{eq:a2}
\end{equation}

\begin{equation}
\sigma^2 = \frac{1}{N}\sum_{i=1}^{N} (Y_i -a_0 - a_1T_i - a_2\hat{Y}_i)^2 \langle \eta_i\rangle \label{eq:sigmma}
\end{equation}

$v$ is the solution of the following non-linear equation.
\begin{equation}
\sum_{i=1}^{N} (\ln{\frac{v}{2}} + 1 + \langle \ln{\eta_i} \rangle - \langle \eta_i \rangle - \Psi(\frac{v}{2})) = 0 \label{eq:v}
\end{equation}






In summary, the inference procedure of the STATE method for count metrics is shown in Algorithm 1.

\begin{algorithm}
\caption{Variational EM Inference with STATE}\label{alg:cap}
\begin{algorithmic}[1]
\State \textbf{Input:} Data$(X_i,T_i,Y_i)_{i=1}^{N}$. 
\State \textbf{Output:} STATE $\hat{a}_1$
\State \textbf{ML stage:} train a model $g(x)$ to predict $E[Y|X]$ by cross-fitting.
\State \textbf{EM stage:} estimate $\hat{a}_1$ as the ATE estimation.
    \begin{displaymath}
          Y_i = a_0 + a_1T_i + a_2\hat{Y}_i + \epsilon_i
    \end{displaymath}
\textbf{initialize:} $(a_0, a_1, a_2, \sigma^2, v)  \gets (a_0^{(0)}, a_1^{(0)}, a_2^{(0)}, \sigma^{2(0)},  v^{(0)})$.
 \While{Free energy $\mathcal{F}$ not converged}
    \State \textbf{E-step} obtain the optimal posterior distribution of $\eta_i$:
    \begin{displaymath}
    q^{(k+1)}(\eta_i) = \mathcal{G}(\eta_i|\xi_i^{(k+1)},\zeta_i^{(k+1)})
    \end{displaymath}
    \State  where $\xi_i^{(k+1)}$,\ $\zeta_i^{(k+1)}$ are computed by Eq.~\eqref{eq:posterior}.
    \State  \textbf{M-step} estimate the model parameters $a_0^{k+1}$, $a_1^{k+1}$, $a_2^{k+1}$, 
    \State $\sigma^{2(k+1)}$, $v^{(k+1)}$ by Eq.~\eqref{eq:a0}-\eqref{eq:v}.

\State  \textbf{Evaluate} $\mathcal{F}$ by Eq.~\eqref{eq:Free_energy}-\eqref{eq:q3}
\EndWhile
\State \textbf{return} STATE: $\hat{a}_1$
\end{algorithmic}
\end{algorithm}


\subsection{Computational Complexity}
Compared to the CUPAC and MLRATE estimators, STATE incurs extra computational costs, primarily due to the implementation of the EM algorithm. Specifically, in each iteration, calculating the posterior distribution parameters $\{\xi_i,\zeta_i\}_{i=1}^{n}$ and model parameters $\theta=\{a^T,\sigma^2,v\}$ each takes $O(N)$ operations. Consequently, the overall computational complexity of EM algorithm amounts to $O(KN)$, where $K$ denotes the number of iterations. According to the empirical experience on the Meituan experimental platform, the EM procedure typically requires 30 to 60 seconds to process 500,000 observations.

\section{Consistent Transformation of Ratio Metrics}
\label{sec:consistent}

The ratio metrics are also common in the industry, which can be regarded as the ratio of two count metrics. For example, the click-through rate is a ratio metric, which is a ratio of the click number and the pageview number. Addressing the variance reduction of ratio metrics is more complex but much less investigated in existing works. 
In this section, we will introduce consistent transformation so that STATE can be utilized to reduce variance of ratio metrics.


Following the notations in Sec.~\ref{subsec:setup}, let $Y_t, Y_c, Z_t, Z_c$ be the corresponding count metrics, i.e., 
$$Y_t = \frac{\sum_{T_i=1} Y_i}{N_t}, Y_c = \frac{\sum_{T_i=0} Y_i}{N_c}, Z_t = \frac{\sum_{T_i=1} Z_i}{N_t}, Z_c = \frac{\sum_{T_i=0} Z_i}{N_c}.$$
Let $R_t$ and $R_c$ be the ratio metrics of the treatment group and the control group respectively, i.e.,
$$R_t = \frac{\sum_{T_i=1} Y_i}{\sum_{T_i=1} Z_i} = \frac{Y_t}{Z_t}, \ R_c = \frac{\sum_{T_i=0} Y_i}{\sum_{T_i=0} Z_i}=\frac{Y_c}{Z_c}.$$
According to the delta method~\cite{dasgupta2008asymptotic}, the variance of ratio metrics can be approximately computed by
\begin{align}
DR_t &= \frac{1}{[EZ_t]^2}DY_t + \frac{[EY_t]^2}{[EZ_t]^4} DZ_t - \frac{2EY_t}{[EZ_t]^3} \mathrm{cov}(Y_t, Z_t) \label{Eq:ratio_variance-1}\\
&= \frac{D[Y_t - \alpha_t Z_t]}{[EZ_t]^2}, \nonumber
\end{align}
where $\alpha_t = EY_t / EZ_t$. Therefore, we have
\begin{align*}
D[\Delta R] &= D[R_t - R_c] \\
&= DR_t + DR_c\\
&= \frac{D[Y_t - \alpha_t Z_t]}{[EZ_t]^2} + \frac{D[Y_c - \alpha_c Z_c]}{[EZ_c]^2},
\end{align*}
where $\alpha_c = EY_c / EZ_c$ and the second equality holds due to $R_t \perp R_c$.

Since the analysis unit in ratio metrics does not match the randomization unit, the common regression-adjusted method and robust estimation cannot directly perform variance reduction for ratio metrics~\cite{Ying2023Optimal}.
The work~\cite{Budylin2018Consistent} transformed ratio metrics $R_t = Y_t/Z_t$ to linear metrics $L_t = Y_t - \kappa Z_t$, where $\kappa = (1 - \eta)R_t + \eta R_c$. After such linear transformation, regression adjustment can be applied for variance reduction of $\Delta L = L_t - L_c$, and the significance level of $\Delta R$ can be obtained by performing student t test for $\Delta L$. However, there are two crucial defects in this work. Firstly, it is showed that $\Delta L = ((1-\eta)Z_c + \eta Z_t) \Delta R$. The coefficient $(1-\eta)Z_c + \eta Z_t$ is not a constant and thus the significance level of $\Delta L$ is not equivalent to that of $\Delta R$. Secondly, the analysis for the variance of $\Delta L$ in this work is based on the hypothesis where the parameter $\kappa$ is taken as a constant, but it actually consists of random variables $R_t$ and $R_c$.

In this section, we follow the main idea in ~\cite{Budylin2018Consistent} (transforming ratio metrics to linear metrics), but mitigate the above two defects simultaneously. Specifically, we construct a new linear metric $\Delta P$ and let $\Delta U = \frac{\Delta P}{EZ_t EZ_c}$, which preserves both unbiased estimation and consistent variance, i.e., 
\begin{enumerate}[\textbf{Property} 1.]
\item Unbiased estimation: $E[\Delta U] = \tau_r,$
\item Consistent variance: $D[\Delta U] \approx D[\Delta R],$
\end{enumerate}
where $\tau_r = \frac{E[Y_i(1)]}{E[Z_i(1)]} - \frac{E[Y_i(0)]}{E[Z_i(0)]}$ is ATE of ratio metrics. Since $EZ_t EZ_c$ is a constant, Property~1 guarantees that the significance level of $\Delta P$ is equivalent to that of $\tau_r$. 
Although the value of $EZ_tEZ_c$ is unknown in practice, we can obtain the significance level of $\tau_r$ by making hypothesis testing for $\Delta P$.
Based on Property~2, 
performing regression adjustment for $\Delta P$ will result in a consistent estimator of $\Delta U$ with smaller variance than $\Delta R$. 
Furthermore, robust estimation can be introduced and eliminate the influence of the atypical and outlying observations.

\textbf{Construction of $\Delta P$.} For each sample, construct the new label as $P_i = \kappa_1 Y_i  - \kappa_2 Z_i$.
Hence, we have $P_t = \frac{\sum_{T_i=1} P_i}{N_t} = \kappa_1 Y_t - \kappa_2 Z_t$, $P_c = \frac{\sum_{T_i=0} P_i}{N_c} = \kappa_1 Y_c - \kappa_2 Z_c$ and $\Delta P = P_t - P_c = \kappa_1 \Delta Y - \kappa_2 \Delta Z$.

\textbf{Proof of Property~1.} Follow the above construction of $\Delta P$ and set $\kappa_1 = Z_c$ and $\kappa_2 = Y_c$. We further have $\Delta P = Y_t Z_c - Z_t Y_c$ and 
\begin{align*}
E[\Delta U] &= \frac{E[Y_t Z_c - Z_t Y_c]}{EZ_tEZ_c} \\
&= \frac{EY_t EZ_c - EZ_t EY_c}{EZ_tEZ_c} \\
&= \frac{EY_t}{EZ_t}  - \frac{EY_c}{EZ_c} \\
&= \frac{E[Y_i(1)]}{E[Z_i(1)]} - \frac{E[Y_i(0)]}{E[Z_i(0)]} \\
&= \tau_r,
\end{align*}
where the second equality holds due to $Y_t,Z_t \perp Y_c, Z_c$. 

\textbf{Proof of Property~2.} Define $g(Y_t, Z_t, Y_c, Z_c) = Y_t Z_c - Z_t Y_c$. According to the delta method, the variance of $\Delta U$ can be rewritten as
\begin{align*}
D[\Delta U] &= \frac{D[Y_t Z_c - Z_t Y_c]}{[EZ_tEZ_c]^2} \\
&= \frac{\nabla^T g(\mu) \sum \nabla g(\mu)}{[EZ_tEZ_c]^2},
\end{align*}
where $\mu = (EY_t, EZ_t, EY_c, EZ_c)$, $\nabla^T g(\mu) = (EZ_c, -EY_c, -EZ_t, EY_t)$ is the Jacobian and 
$$\sum = \begin{pmatrix}
DY_t & \mathrm{cov}(Y_t,Z_t) & 0 & 0\\
\mathrm{cov}(Y_t,Z_t) & DZ_t & 0 & 0\\
0 & 0 & DY_c & \mathrm{cov}(Y_c,Z_c)\\
0 & 0 & \mathrm{cov}(Y_c,Z_c) & DZ_c
\end{pmatrix}$$
is the covariance matrix.
Therefore, we further have
\begin{align*}
D[\Delta U] =& \frac{\nabla^T g(\mu) \sum \nabla g(\mu)}{[EZ_tEZ_c]^2} \\
=& \frac{1}{[EZ_t]^2} \left [ DY_t + \frac{[EY_c]^2}{[EZ_c]^2} DZ_t - \frac{2EY_c}{EZ_c} \mathrm{cov}(Y_t, Z_t) \right ] + \\
& \frac{1}{[EZ_c]^2} \left [ DY_c + \frac{[EY_t]^2}{[EZ_t]^2} DZ_c - \frac{2EY_t}{EZ_t} \mathrm{cov}(Y_c, Z_c) \right ] \\
=& \frac{D[Y_t - \alpha_c Z_t]}{[EZ_t]^2} + \frac{D[Y_c - \alpha_t Z_c]}{[EZ_c]^2},
\end{align*}
where $\alpha_t = EY_t / EZ_t$ and $\alpha_c = EY_c / EZ_c$. As stated in~\cite{Budylin2018Consistent}, in the real online experiments of the industry, the increment in ratio metrics of the treatment group is usually small and is tightly bounded with deviation of several percents. Hence, we have $\alpha_t \approx \alpha_c$, which indicates that $D[\Delta U] \approx D[\Delta R]$ holds.

\section{Experimental Evaluation}
In this section, we construct large-scale experiments to demonstrate the effectiveness of STATE using both simulated data and real user data from the Meituan food delivery platform. Firstly, we validate the performance of various methods in reducing the variance of count metrics and ratio metrics in simulation data. For completeness, we also investigate the impact of the proportion of outliers in the data set on STATE. Secondly, to show the magnitude of variance reduction that can be achieved in practice, we perform a series of A/A tests on key metrics of interest for Meituan food delivery business. Finally, we discuss limitations of the STATE estimator.

\subsection{Experimental Setup}

\textbf{Simulation data.} The generating process is almost the same as ~\cite{Guo2021Machine}. Denote each sample by $(X_i, T_i, Y_i, Z_i)$. The covariates $X_i$ is distributed as $X_i \sim \mathcal{N}(u, I_{5 \times 5})$, where $u$ is generated from a uniform distribution $U(0,10)$. 
The outcome variables $Y_i$ and $Z_i$ are constructed by $Y_i = b(X_i) + T_i * \tau_y(X_i) + \epsilon_i$ and $Z_i = c(X_i) + T_i * \tau_z(X_i) + \eta_i$ respectively, where $\epsilon_i$ and $\eta_i$ are the error terms distributed as $\epsilon_i \sim \mathcal{N}(0, 25^2)$ , $\eta_i \sim \mathcal{N}(0, 10^2)$.  Let $b(X_i)$ and $c(X_i)$ be nonlinear functions with the forms  
$$
\ \ \ \ \ \ \ \ \ \left \{
\begin{aligned}
b(X_i) &= 10 \sin(\pi X_{i1}X_{i2}) + 6X_{i3}^2 + 10|X_{i4}| + 5|X_{i5}| + 50, \\
c(X_i) &= 10 \sin(\pi X_{i4}) X_{i5} + 15(X_{i2} + X_{i3})^2 + 5|X_{i1}| + 30.
\end{aligned}
\right .
$$
The treatment effect is also given by a nonlinear form as 
$$
\ \ \ \ \ \ \ \ \ \left \{
\begin{aligned}
\tau_y(X_i) &= X_{i1} X_{i3} + \log (1 + e^{X_{i2}}), \\
\tau_z(X_i) &= X_{i2}^2 + 3\log(1 + e^{X_{i4}} + |X_{i5}|).
\end{aligned}
\right .
$$
Additionally, in order to validate the performance of our STATE estimator in dealing with heavy-tailed metrics, we also add a certain proportion of outliers to the simulation data set.
The outliers are randomly generated from two uniform distributions $U(\overline{Y} + 4 \sigma_y,  \overline{Y} + 20 \sigma_y)$ and $U(\overline{Z} + 4 \sigma_z, \overline{Z} + 20 \sigma_z)$, where $\sigma_y$ and $\sigma_z$ are the standard deviation of $Y_i$ and $Z_i$ respectively.

We generate 200K samples in total and perform 1000 simulation experiments. In each simulation, we randomly choose 20K samples and divide them equally into the treatment group and the control group according to the treatment indicator $T_i$ which follows $T_i \sim \mathrm{Bernoulli}(0.5)$. 
Finally, we compare the variance of different ATE estimators in all the simulations.

\noindent\textbf{Real user data.} The real user data is collected from Meituan food delivery platform, which contains over 2.3 million samples and 58 covariates. Each sample records all the transaction information for a user during an experiment in this platform. To show the effectiveness of the STATE estimator, we select two count metrics, the count of orders per user (orders in short) and GMV per user, and a ratio metric, the average price per order (AvgOrdPrice). Here, GMV refers to the total price paid by each user on the food delivery platform during the trial period. 
The average price per order is calculated by dividing the sum of GMV of all users by the total number of orders of all users.

For each metric, we perform A/A tests by selecting 200k users randomly and assigning the treatment indicator $T_i \sim \mathrm{Bernoulli}(0.5)$ for each user. Similarly, the A/A test is also repeated 1000 times. Notice that the A/A test is a controlled experiment where treatment is identical to control, hence the ground truth of ATE is 0.

\noindent\textbf{Benchmark.} For both count and ratio metrics, multiple methods for variance reduction are implemented and taken as the benchmarks.
\begin{itemize}
\item Count metrics
\begin{itemize}
\item DIM. The difference-in-mean estimator that computes ATE by Eq.~\eqref{Eq:Count-DIM} for count metrics.
\item CUPED. The state-of-the-art linear method proposed in~\cite{Deng2013Improving}, which reduces variance of count metrics by utilizing the pre-experiment covariates and gives an unbiased estimation of ATE. 
\item CUPAC. Similar to CUPED, but replace the pre-experiment covariates with a proxy highly correlated with the outcome variable~\cite{Huang2020Control}.
\item MLRATE. The machine learning regression-adjusted estimator proposed in~\cite{Guo2021Machine}.
\item STATE. The robust regression-adjusted estimator for variance reduction proposed in this paper.
\end{itemize}
\item Ratio metrics
\begin{itemize}
\item DIM. The difference-in-mean estimator that computes ATE of ratio metrics by Eq.~\eqref{Eq:Ratio-DIM}.
\item CUPED. The generalization to ratio metrics proposed in Appendix~B of~\cite{Deng2013Improving}. 
\item CTRM. The consistent transformation method of ratio metrics proposed in~\cite{Budylin2018Consistent}.  
\item STATE. The robust regression-adjusted estimator for ratio metrics after consistent transformation.
\end{itemize}
\end{itemize}

\noindent\textbf{Evaluation Metrics.} We compare various variance reduction techniques on two primary evaluation metrics: bias and variance. Consistent with previous studies\cite{Guo2021Machine,Ying2023Optimal}, we assess the bias by utilizing the empirical coverage rate of the 95\% confidence intervals (CI).  
The closer the empirical coverage rate is to the nominal coverage rate(95\%), the smaller the bias of the estimator. 
On the other hand, for the variance metric, the smaller the value, the shorter the confidence interval and the higher the efficiency of the estimator.

\begin{itemize}
\item Empirical coverage. The empirical coverage represents the proportion of 1000 simulations in which the 95\% confidence interval covers the ground truth. 
\item Variance. The variance of ATE estimates across 1000 simulation experiments.
\end{itemize}

\subsection{Simulation Experiment}
\begin{figure}[bp]
\centering  
\subfigure[PDF of ATE for count metric]{
\label{fig:count_kde}
\includegraphics[width=0.22\textwidth]{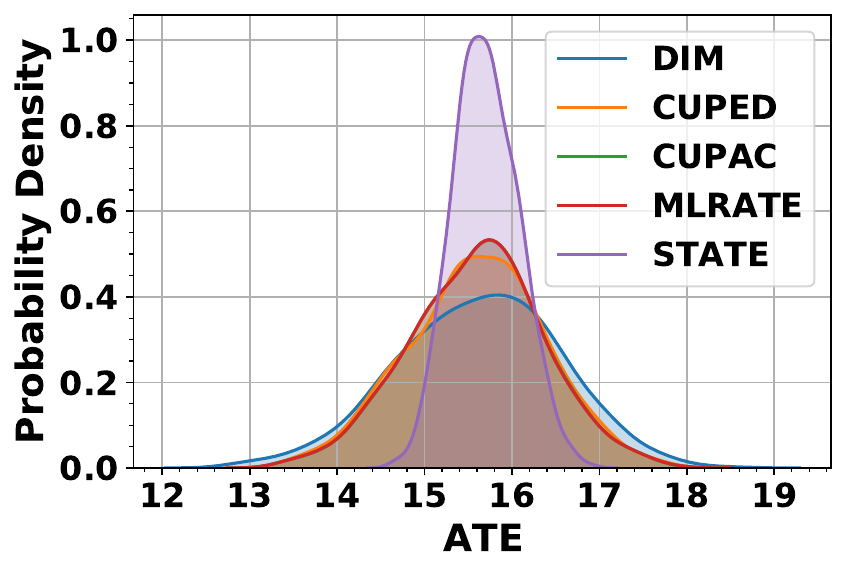}}
\subfigure[PDF of ATE for ratio metric]{
\label{fig:ratio_kde}
\includegraphics[width=0.225\textwidth]{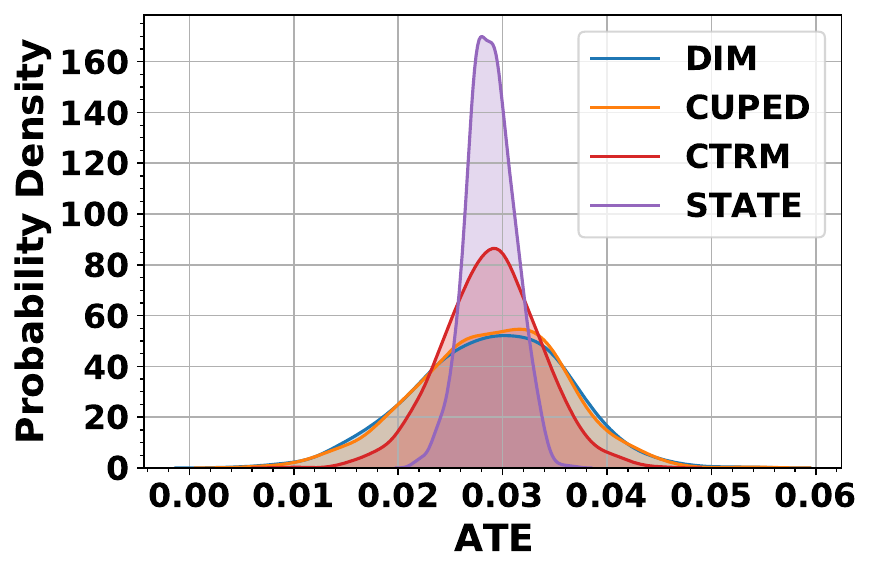}}
\caption{Simulation results of ATE estimation}
\label{fig:Simulation_kde}
\end{figure}

In this section, we utilize simulation experiments to validate the effectiveness of the algorithms proposed in this paper. First of all, Table~\ref{tab:count_simulation}, ~\ref{tab:ratio_simulation} and Fig.~\ref{fig:Simulation_kde} show the simulation results for count metrics and ratio metrics on the dataset with 0.5\% outliers. Furthermore, we investigate the impact of the proportion of outliers on various variance reduction techniques. "Var.Red\%" displays the variance reduction gains for each method compared to DIM estimator. "Emp.Cov\%" displays the empirical coverage of the 95\% CIs.

\begin{table}[htbp]
    \centering
    \caption{Simulation Results for count metric}
    \label{tab:count_simulation}  
\begin{tabular}{*{6}{c}}
  \hline
  Methods     & DIM  & CUPED & CUPAC & MLRATE & STATE \\ 
  \hline
  Emp.Cov\%  & 94.6 & 94.6  & 94.2  & 94.0 &   95.6  \\
  Var.Red\%  & 0    & 28.7  & 32.6  & 32.4 &   84.1  \\
  \hline
\end{tabular}
\end{table}

\begin{table}[htbp]
    \centering
    \caption{Simulation results for ratio metric}
    \label{tab:ratio_simulation}  
\begin{tabular}{*{5}{c}}
  \hline
  Methods     & DIM  & CUPED & CTRM  & STATE \\ 
  \hline
  Emp.Cov\%  & 95.1 & 94.3  & 94.6   &   94.8  \\
  Var.Red\%  & 0    & 6.7  & 55.7    &   90.1  \\
  \hline
\end{tabular}
\end{table}


\subsubsection{Count Metrics}
\ 

\vspace{1mm}
As shown in Table~\ref{tab:count_simulation}, the empirical coverage of all count metric estimators closely approximates the nominal coverage (95\%), indicating that the biases associated with these estimators are minimal.

Fig.~\ref{fig:count_kde} depicts the probability density of ATE estimates, revealing that STATE's ATE distribution is markedly more compact, with its variance being substantially reduced compared to that of other methods. Specifically, the CUPED method reduces the variance by 28.7\% compared to the DIM estimator; CUPAC and MLRATE perform similarly and deliver additional gains relative to CUPED. It indicates that the proxy variable constructed by the ML model has a superior correlation with the outcome than a single covariate. 
In contrast, the performance of the STATE estimator is remarkable, achieving an over 80\% variance reduction compared to the DIM estimator and significantly enhancing the precision of ATE estimation. 
Considering both the empirical coverage rate and estimator variance, STATE clearly emerges as the preferred ATE estimator.

\subsubsection{Ratio Metrics}
\

\vspace{1mm}
Fig.~\ref{fig:ratio_kde} and Table~\ref{tab:ratio_simulation} present the detailed experimental results of variance reduction for ratio metrics in simulation data. As is shown by Table~\ref{tab:ratio_simulation}, the empirical coverage of all the estimators are close to the nominal coverage, in which STATE achieves the best performance on variance reduction. Specifically, CUPED only reduces 6.7\% of variance compared to DIM duo to the weak correlation between the pre-experiment covariates and the ratio metric. CTRM performs much better, which achieves over 55.7\% variance reduction compared with DIM. However, the effectiveness of CTRM is also limited by the correlation between the ML-based predictors and the outcome metrics, which is easily influenced by big outliers. 
As a robust estimator, STATE can significantly decrease the negative effects for variance caused by outliers, which contributes to over 90.1\% variance reduction relative to DIM.

\begin{table*}[htbp]
    \centering
    \caption{Summary of A/A test results for count metrics}
    \label{tab:count_AA}  
\begin{tabular}{*{12}{c}}
  \toprule
  \multirow{2}*{Metric} & \multicolumn{1}{c}{DIM} & \multicolumn{2}{c}{CUPED} & \multicolumn{2}{c}{CUPAC} & \multicolumn{2}{c}{MLRATE} & \multicolumn{2}{c}{STATE} \\ 
  \cmidrule(lr){2-2} \cmidrule(lr){3-4} \cmidrule(lr){5-6} \cmidrule(lr){7-8} \cmidrule(lr){9-10} 
  & Emp.Cov\%  & Var.Red\% & Emp.Cov\% & Var.Red\% & Emp.Cov\% & Var.Red\% & Emp.Cov\% & Var.Red\% & Emp.Cov\% \\
  \midrule
  orders  &  94.9 & 47.2 & 94.9 & 54.6 & 94.5 & 54.6 & 94.5 & 70.5 & 95.1\\
  GMV     &  94.8 & 33.3 & 93.9 & 44.3 & 94.5 & 44.4 & 94.4 & 80.7 & 95.2\\
  \bottomrule
\end{tabular}
\end{table*}

\begin{table*}[htbp]
    \centering
    \caption{Summary of winsorized results at the 99.9th percentile threshold}
    \label{tab:count_AA_Winsorized_99}  
\begin{tabular}{*{12}{c}}
  \toprule
  \multirow{2}*{Metric} & \multicolumn{2}{c}{Winsorized DIM} & \multicolumn{2}{c}{Winsorized CUPED} & \multicolumn{2}{c}{Winsorized CUPAC} & \multicolumn{2}{c}{Winsorized MLRATE} & \multicolumn{2}{c}{Huber Regression} \\ 
  \cmidrule(lr){2-3} \cmidrule(lr){4-5} \cmidrule(lr){6-7} \cmidrule(lr){8-9} \cmidrule(lr){10-11} 
  &Var.Red\% & Emp.Cov\%  & Var.Red\% & Emp.Cov\% & Var.Red\% & Emp.Cov\% & Var.Red\% & Emp.Cov\% & Var.Red\% & Emp.Cov\% \\
  \midrule
  orders & 1.1  &  94.7 & 49.0 & 94.5 & 55.9 & 94.4 & 55.9 & 94.3 & 54.9 & 95.0\\
  GMV    & 12.4 &  94.7 & 38.3 & 93.9 & 57.3 & 94.3 & 54.6 & 94.4 & 47.3 & 94.5\\
  \bottomrule
\end{tabular}
\end{table*}

\begin{table*}[htbp]
    \centering
    \caption{Summary of winsorized results at the 99th percentile threshold}
    \label{tab:count_AA_Winsorized_999}  
\begin{tabular}{*{12}{c}}
  \toprule
  \multirow{2}*{Metric} & \multicolumn{2}{c}{Winsorized DIM} & \multicolumn{2}{c}{Winsorized CUPED} & \multicolumn{2}{c}{Winsorized CUPAC} & \multicolumn{2}{c}{Winsorized MLRATE} & \multicolumn{2}{c}{Huber Regression}\\ 
  \cmidrule(lr){2-3} \cmidrule(lr){4-5} \cmidrule(lr){6-7} \cmidrule(lr){8-9} \cmidrule(lr){10-11} 
  &Var.Red\% & Emp.Cov\%  & Var.Red\% & Emp.Cov\% & Var.Red\% & Emp.Cov\% & Var.Red\% & Emp.Cov\% & Var.Red\% & Emp.Cov\%\\
  \midrule
  orders & 9.76  &  94.1 & 53.0 & 94.0 & 60.4 & 94.1 & 60.5 & 94.0 & 54.9 & 95.0\\
  GMV    & 21.7  &  94.5 & 49.8 & 93.5 & 71.0 & 94.0 & 67.7 & 93.8 & 47.3 & 94.5\\
  \bottomrule
\end{tabular}
\end{table*}

\subsubsection{Factors Affecting STATE Effectiveness}
\

\begin{figure}[htbp]
\centering  
\subfigure[Empirical coverage for all estimators]{
\label{fig:Emp_Cov}
\includegraphics[width=0.225\textwidth]{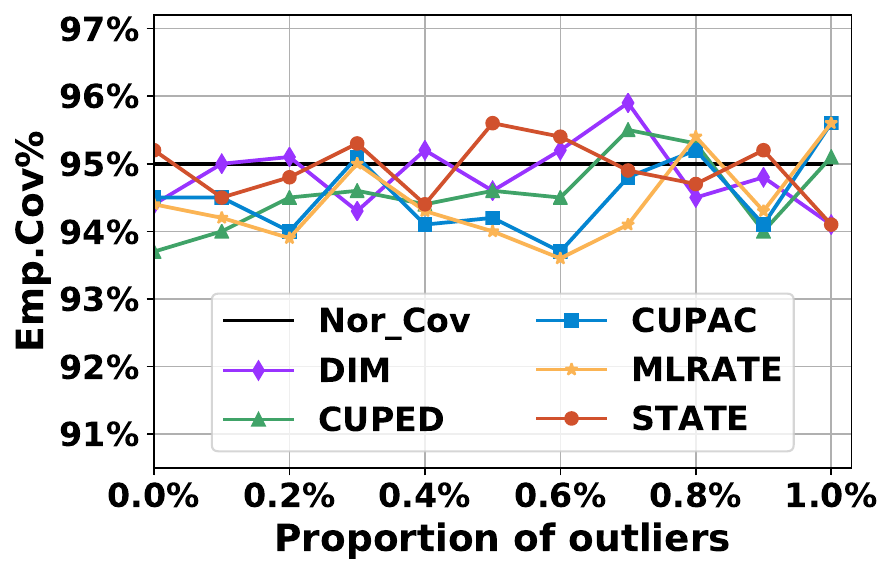}}
\subfigure[Variance reduction compared to DIM]{
\label{fig:Var_Red}
\includegraphics[width=0.23\textwidth]{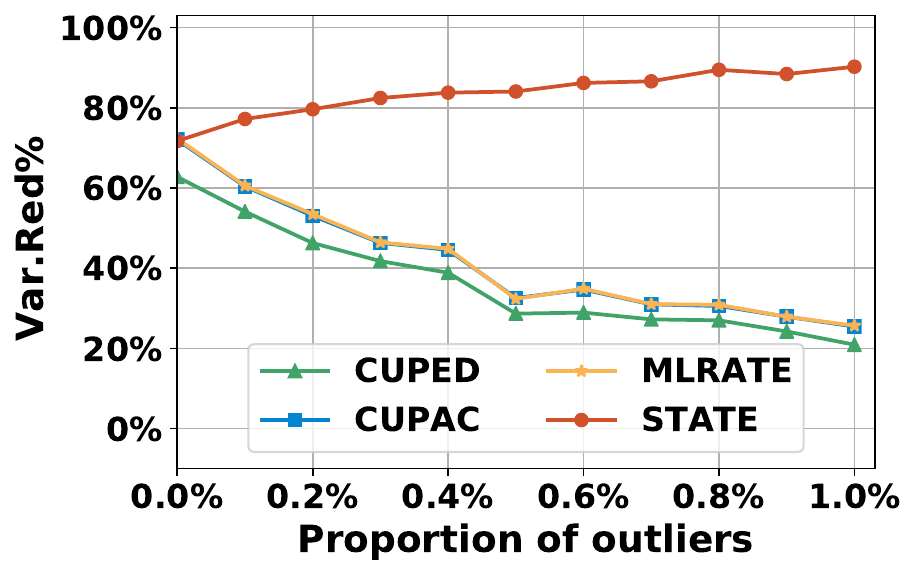}}
\caption{Impact of proportions of outliers}
\label{fig:outliers}
\end{figure}

\vspace{1mm}
Now let's look at the factors that affect the performance of STATE. Here, we discuss the problem from a general setting to specific settings by constructing a Gaussian distribution dataset and adding different proportions of outliers from 0 to 1\%.
Fig.~\ref{fig:Emp_Cov} and Fig.~\ref{fig:Var_Red} display the results of empirical coverage and variance reduction respectively.

From the Fig.~\ref{fig:Var_Red}, we can see that when the dataset does not contain outliers, more precisely, the metrics follow a Gaussian distribution, STATE, CUPAC, and MLRATE perform similarly in variance reduction.
As the proportion of outliers increases, STATE and baseline methods exhibit drastically different performance. The reason is that the degree of variance reduction of typical methods such as CUPED, CUPAC and MLRATE depends on the correlation between the constructed covariates or proxy variables and business metrics. However, with the existence of outliers, correlation learning becomes increasingly difficult. The result is that the performance of these typical techniques deteriorates gradually, degrading to the variance scale of the DIM method. The STATE estimator we proposed is designed to address such challenges by modeling the noise as a Student's $t$-distribution. Because of the excellent robustness of the T estimate, the rate of variance increase of STATE is significantly slower than the DIM method with the increase of outliers. Consequently, the more pronounced the heavy tail phenomenon of business metrics, the more apparent the advantage of the STATE method in variance reduction.



\begin{table*}[htbp]
    \centering
    \caption{Summary of A/A test results for ratio metrics}
    \label{tab:ratio}  
\begin{tabular}{*{8}{c}}
  \toprule
  \multirow{2}*{Metric} & \multicolumn{1}{c}{DIM} & \multicolumn{2}{c}{CUPED} & \multicolumn{2}{c}{CTRM}   & \multicolumn{2}{c}{STATE} \\ 
  \cmidrule(lr){2-2} \cmidrule(lr){3-4} \cmidrule(lr){5-6} \cmidrule(lr){7-8} 
  & Emp.Cov\% & Var.Red\% & Emp.Cov\%  & Var.Red\% & Emp.Cov\% & Var.Red\% & Emp.Cov\%  \\
  \midrule
  AvgOrdPrice &  95.0 & 6.4 & 95.6 & 21.8 & 94.9  & 91.6 & 95.7 \\
  \bottomrule
\end{tabular}
\end{table*}

\subsection{Empirical Results }
In this section, we perform two real experiments on the Meituan food delivery platform, analyzing the count metrics and ratio metrics, respectively. Here, we focus on the A/A tests, not the A/B tests running in production. This is because the true effect of the A/B tests is unknown, which makes it impossible to evaluate the empirical coverage rate of the confidence interval. Since the average treatment effect (ATE) in online experiments is typically small and unlikely to significantly change the relationship between the outcomes and covariates, the reduction in variance in A/B tests should be very similar to that in A/A tests.

\begin{figure}[htbp]
\centering  
\subfigure[PDF of ATE for orders]{
\label{fig:order_kde}
\includegraphics[width=0.237\textwidth]{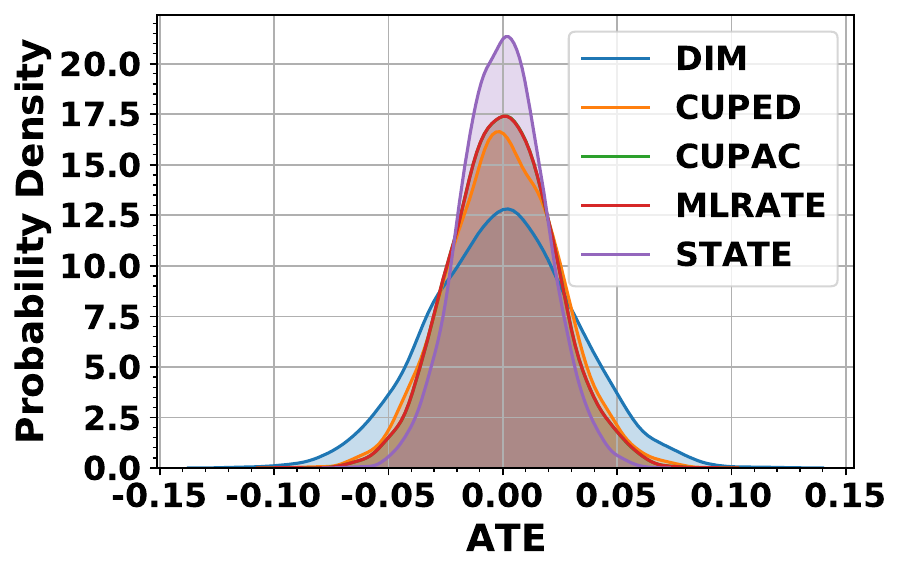}}
\subfigure[PDF of ATE for GMV]{
\label{fig:total_kde}
\includegraphics[width=0.225\textwidth]{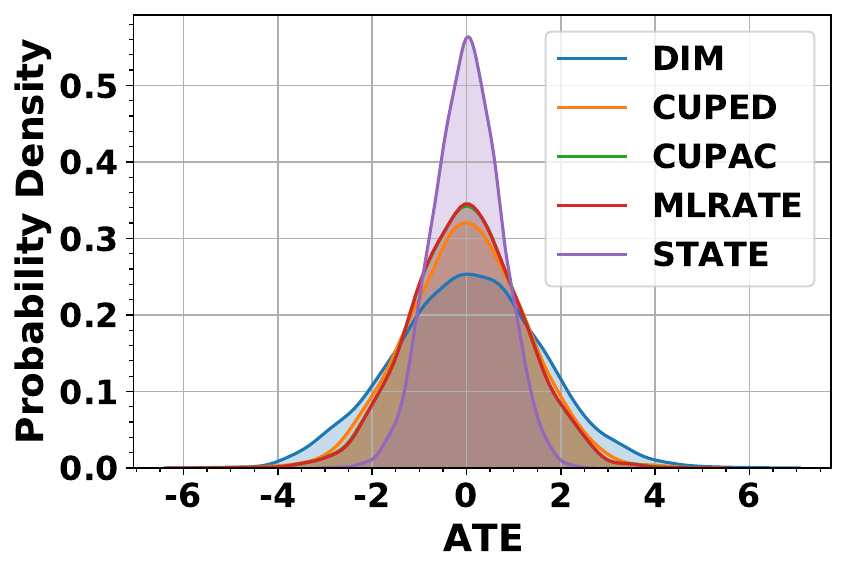}}
\subfigure[PDF of ATE for AvgOrdPrice]{
\label{fig:AvgOrdPrice_kde}
\includegraphics[width=0.225\textwidth]{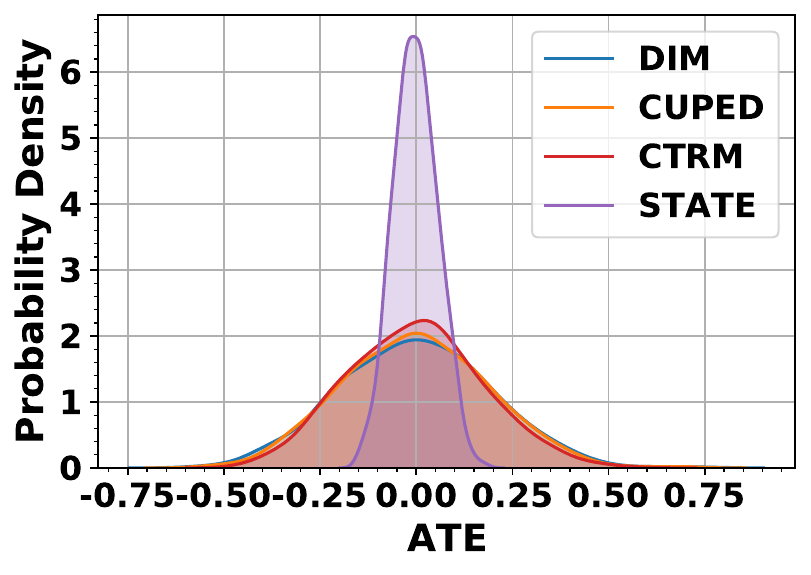}}
\caption{Empirical results of ATE estimation}
\label{fig:Empirical_kde}
\end{figure}

\subsubsection{Count metrics}

\ 

\vspace{1mm}
On the food delivery platform, business metrics with a heavy-tailed distribution are quiet common. This is typically caused by users' accidental consumption behavior. For example, a user orders takeaway to treat friends to dinner, the consumption price of her order may be a extremely large value relative to her historical consumption pattern. 
These sample points that fall on the tail (also known as outliers) are not affected by treatment, but has great impact on the sensitivity of measuring ATE. 

To get a better sense of the magnitudes of variance reduction that the STATE method might achieve in practice, we select two key count metrics: orders per user and GMV per user, both of which have heavy-tail distributions. For each metric, we construct A/A tests. Table~\ref{tab:count_AA} and Fig.~\ref{fig:Empirical_kde} show the empirical results. 
It indicates that the STATE performs substantially better than the CUPED, CUPAC and MLRATE estimators on both orders metric and GMV metric. On average, the STATE estimator reduces the variance of the order metric and GMV metric by 70.5\% and 80.7\% respectively compared to the DIM method, whereas the analogous figures for the state-of-the-art methods are about 54.6\% and 44.4\%. It demonstrates that STATE can indeed considerably improve the sensitivity of real business metrics, which is of great significance for increasing business profits and reducing the cost of experimental time. 

Furthermore, we compare STATE method with the typical winsorization ~\cite{dixon1960simplified} method and Huber regression~\cite{burke2019measuring} for dealing with heavy-tailed metrics. Table~\ref{tab:count_AA_Winsorized_99} and ~\ref{tab:count_AA_Winsorized_999} display the results of winsorization at different thresholds(99\% and 99.9\% quantiles of the observations). It is observed that estimation is sensitive to the threshold. A lower threshold results in more observations beyond threshold being restrained, which consequently yields a reduced variance in parameter estimation. However, there is a concomitant decline in the empirical coverage rate, suggesting an incremental increase in bias. Consequently, employing the Winsorization process necessitates a meticulous balance between the impact of bias and variance. 
As shown in Table~\ref{tab:count_AA} and~\ref{tab:count_AA_Winsorized_99}, the variance of the parameters estimated by the Huber regression method is slightly smaller than that of the CUPAC and MLRATE methods (without winsorization). This is because Huber regression transforms the mean square loss function of extreme observations into a linear form, thereby partially reducing the impact of outliers. However, the results also show that the robustness of the Huber regression method is limited, exhibiting a considerable disparity when compared to the STATE method.


\subsubsection{Ratio metrics}

\ 

\vspace{1mm}

Fig.~\ref{fig:AvgOrdPrice_kde} and Table~\ref{tab:ratio} present the results for ratio metric named AvgOrdPrice in real user data, which is calculated by dividing the
sum of GMV of all users by the total number of orders. 
As is shown by Table~\ref{tab:ratio}, the empirical coverage of all the estimators converges
to 95\%, and STATE still performs the best because of the strong capacity of resisting disturbance for outliers that exactly exists in real data. In summary, the variance reduction of STATE is about 91.6\% relative to DIM, relative to CUPED is 91.0\%, relative to CTRM is 89.3\%.

\subsection{Limitations}
In the preceding discussions, STATE has achieved notable success in the application of variance reduction for both count metrics and ratio metrics. Now let's discuss the applicability of the STATE method. Inspired by section 5.2.3, we find the effectiveness of the STATE method varies from metric to metric, depending on whether the distribution of the metric has a heavy tail. When metric follows a Gaussian distribution, the STATE method is comparable with the state-of-the-art methods, but introduces extra computational cost. If the metric exhibits a heavy tail distribution, the STATE method demonstrates a strong advantage.

\section{Conclusion}

Variance reduction is the common technology to improve the sensitivity in online controlled experiments, which contributes to smaller user population and shorter experimental period. However, typical methods cannot characterize the heavy-tailed distributions in real business metrics, whose efficiency is greatly limited by the outlying observations.
In this paper, we proposed a machine-learning-based regression adjustment method with Student's $t$-distributed errors to reduce the impact of outliers in variance reduction for count metrics. 
Furthermore, we transform ratio metrics into a linear combination but preserve unbiased estimation and consistent variance, and apply the above method to the variance reduction of ratio metrics. Both synthetic and real data demonstrate a significant decrease in variance for both count and ratio metrics compared to state-of-the-art methods. 
It benefits from our robust estimator that is stronger to resist disturbance of outliers. 
We recommend to use the method especially in the experiments where the treatment effect is not very significant.
In this case, the influence of outliers is relative larger but can be effectively mitigated by our method.

As the method significantly improves the robustness of controlled experiment results, we would like to expand it to the estimation of individual treatment effects (ITE) in future works, which also suffers from the negative effects of outliers.

\begin{acks}
This work was supported in part by National Key R\&D Program of China (2022YFB290180), the NSF of China (62172206), and the Xiaomi Foundation.
\end{acks}

\bibliographystyle{ACM-Reference-Format}
\bibliography{reference}











\end{document}